\documentclass[preprint, 12pt]{elsarticle}

\journal{TBD}

\usepackage{graphicx}
\usepackage{amssymb}
\usepackage{amsmath}
\usepackage{amsfonts}
\usepackage{listings}
\usepackage{url}
\usepackage{lineno}

\usepackage{natbib}
\usepackage[noabbrev,capitalize]{cleveref}
\usepackage[dvipsnames]{xcolor}

\usepackage{gensymb}

\newcommand{\mycomment}[1]{}

\begin{document}

\begin{frontmatter}


\title{Assessing model error in counterfactual worlds}

\author[eah]{Emily Howerton}%\orcidlink{0000-0002-0639-3728}}
\ead{ehowerton@princeton.edu}
\address[eah]{Department of Ecology and Evolutionary Biology, Princeton University}

\author[jtl]{Justin Lessler}%\orcidlink{0000-0002-9741-8109}}
\ead{jlessler@unc.edu}
\address[jtl]{Department of Epidemiology, University of North Carolina at Chapel Hill}

\begin{abstract}
Counterfactual scenario modeling exercises that ask ``what would happen if?"  are one of the most common ways we plan for the future. Despite their ubiquity in planning and decision making, scenario projections are rarely evaluated retrospectively. Differences between projections and observations come from two sources: scenario deviation and model miscalibration. We argue the latter is most important for assessing the value of models in decision making, but requires estimating model error in counterfactual worlds.  Here we present and contrast three approaches for estimating this  error, and demonstrate the benefits and limitations of each in a simulation experiment. We provide recommendations for the estimation of counterfactual error and discuss the components of scenario design that are required to make scenario projections evaluable.

%make it difficult to tell the extent to which a model was miscalibrated. Yet, retrospective evaluation is essential for building trust and improving the models that generate projections. 
%Here we present  and compare three methods for estimating the error in model projections in counterfactual worlds. We further show how, using these estimates, observed error can be decomposed into its two sources: model miscalibration and scenario deviation. 
%Here, we present a theoretical framework for evaluating scenario projections. We show how two sources of error,  scenario deviation and model miscalibration, can interact, and we present three approaches for teasing them apart. We demonstrate the benefits and limitations of each approach in simulations

%. Finally, we discuss the important components of scenario design that are required to make a scenario projection evaluable.

\end{abstract}


\end{frontmatter}

% \linenumbers

\section{Introduction}
\label{S:1}

Predictive models are a central tool in many fields for assessing the state of the world and planning for the future. Models enable us to not only forecast \textit{what will happen} but also
to predict \textit{what would happen if}  certain conditions were to occur. For instance, what if we limit global surface warming to 2\degree C? Or, what if 80\% of seniors received a vaccination against seasonal influenza? By helping us to determine our best possible answer to these questions given current knowledge, models play an essential role in managing risk and informing decisions. 

These \textit{what would happen if} scenario projections are  most useful when we are interested in the effects of decisions or key uncertainties \cite{runge_scenario_2024}. When we aim to intervene in the present to change outcomes in the future, scenario projections can help quantify the benefits and costs of potential actions. If key drivers of dynamics are highly uncertain, scenario projections can help define future outcomes under different assumptions (e.g., best and worst cases). In many situations we are making decisions under uncertainty, hence are simultaneously interested in both. 

% Scenario projections are used in many fields, including perhaps the most famous CMIPP climate projections as well as projections of infectious disease outcomes more recently popularized during the COVID-19 pandemic. \textcolor{red}{Details here} 

Although predictions of the future are typically generated to inform planning and response in the present, retrospective evaluation of predictions is  essential to improving predictive models \cite{alley_advances_2019} and building trust with end users \cite{raftery_use_2016}. There is well-developed theory and methodology for evaluating forecasts \cite{gneiting_probabilistic_2014}, where forecasts of \textit{what will happen} are compared to observations that eventually materialize.  The goal of this assessment is to evaluate whether forecasts are ``well calibrated", or consistent with  observations. 
%For probabilistic predictions, the uncertainty, or ``sharpness" is also considered. Typically, the goal is to ``maximize sharpness subject to calibration", as sharper, more certain predictions are more informative, insofar as they are accurate \cite{gneiting_probabilistic_2014, gneiting_probabilistic_2007}.  

These evaluation metrics, however, cannot be straightforwardly applied to scenario projections. When we make projections under multiple scenarios, at most one set of scenario assumptions will be true, and likely none will hold exactly. We cannot directly evaluate the calibration of projections for these counterfactual worlds that we never observe. For example, an airline might project annual earnings under scenarios of a 2\% and 5\%  increase in travel activity, but in actuality see a 3\% increase for which they had no projection. Because of this fundamental challenge, scenario projections are rarely evaluated in practice. However, without evaluation, biases in the models can go unidentified and mislead decision making.

Climate modelers have long grappled with this challenge. Because the scenarios that were assumed cannot be directly evaluated against observed data, they have instead proposed a range of other metrics for quality assessment and trust building. These include, for example, the ability of a model to reliably recreate past conditions or its consistency with physical laws \cite{tebaldi_use_2007}. More recently, a few attempts to directly evaluate scenario projections have been made in climate \cite{hausfather_evaluating_2020} and for infectious diseases \cite{howerton_evaluation_2023}. However, as we will discuss below, these methods each have limitations.

Here, we present a conceptual framework for evaluating scenario projections. We describe the central challenge with scenario projection evaluation, situate existing methods within our conceptual framework, and propose three general approaches to overcome these challenges, each with their own requirements and assumptions. Finally, we demonstrate the feasibility and limitations of each approach in a simulation experiment. 

\section{The challenge with evaluating scenario projections: Two sources of error}
\label{S:2}

The fundamental challenge with evaluating the calibration of scenario projections arises from their counterfactual nature. Because a scenario projection specifies \textit{what would happen if} a particular set of conditions were to be realized, the explicit comparison of a projection to an observation (as in forecast evaluation) is difficult to interpret because there are two potential sources contributing to the error: scenario deviation and model miscalibration. 

\begin{figure}[!htbp]
\centering
\includegraphics[width=0.9\linewidth]{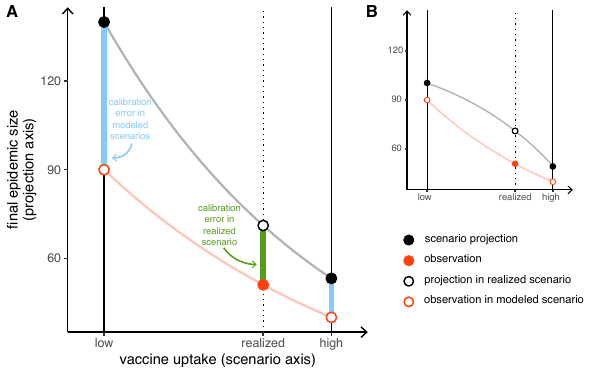}
\caption{Scenario projection schematic. (A) For low vaccination and high vaccination uptake scenarios, projections for final epidemic size, $(x_i, \mathcal{P}(y|x_i))$, are shown (black solid points). An observation consisting of a realized vaccination uptake and corresponding epidemic size, $(x^*, \mathcal{P}^*(y|x^*))$, is realized after the projection period (red solid point). Some relationship between projected outcomes along the vaccination exists for the model (gray line) and the truth (red line), but they likely will not be available in practice. A projection for the realized scenario, $\mathcal{P}(y|x^*)$, can be obtained retrospectively (open black point) if this relationship can be estimated or if the models can be rerun retrospectively. For each of the modeled scenarios, an observation exists conceptually, $\mathcal{P}^*(y|x_i)$, but will never be observed (open red points). We can obtain the \textit{calibration error in the realized scenarios} (green bar) by comparing the observation to the model projection from the realized scenario, $\mathcal{P}^*(y|x_i) - \mathcal{P}^*(y|x^*)$. But, we are ultimately interested in \textit{calibration error in the modeled scenarios} (blue bars), $\mathcal{P}(y|x_i) - \mathcal{P}^*(y|x_i)$. (B) A second set of projections, with a different relationship along the scenario axis. Projections in (A) and (B) have equivalent error in the realized scenario, but different error in the modeled scenarios.}
\label{fig:base-plot}
\end{figure}

To illustrate this fundamental challenge, we present a conceptual schematic of a scenario projection (\cref{fig:base-plot}). Let $\mathcal{P}^{m}(y|x_i)$ be a projection of outcome $y$ from model $m$ under conditions $x_i$ specified along scenario axis $x$ for scenario $i$. For example, consider a case where low and high vaccination coverage scenarios are of interest (x-axis, \cref{fig:base-plot}), and a model generates projections of cumulative hospitalizations for each of these assumptions over a 6-month period (black points, \cref{fig:base-plot}). After the projection period has passed, we observe the realized vaccination coverage, $x^*$, and the corresponding realized cumulative hospitalizations, $y^* = \mathcal{P}^*(y|x^*)$ (red point, \cref{fig:base-plot}).  

In a forecasting setting, $x^*$ is irrelevant because the forecast of \textit{what will happen} is expected to encompass all possible future values of $x$ (and, implicitly, their likelihood). However, for a scenario projection, which is conditional on $x_i$, $x^*$ will match at best only one of the scenarios that were modeled, and usually $x^*$ will not match any of the scenarios modeled. Despite this mismatch, it might be tempting to compare the projection $\mathcal{P}^{m}(y|x_i)$ and the observation $y^*$ directly. For example, if scenario projections and observations are plotted together, this is the natural comparison. Importantly, though, this comparison is difficult to interpret because there are two potential drivers of differences between the projected values and observation (\cref{fig:base-plot}): 

\begin{itemize}
    \item \textbf{scenario deviation:} the world did not unfold according to the scenarios modeled (i.e., $x_i \neq x^*$). Thus, projected values from any scenario, $\mathcal{P}^{m}(y|x_i)$, should not necessarily match observed values from the ``realized" scenario, $\mathcal{P}^*(y|x^*)$. 

    \item \textbf{model miscalibration:} even if the projections had been made for the perfect scenario (i.e., $x_i = x^*$, white circle \cref{fig:base-plot}), the projection may still deviate from observations. That is, model miscalibration is the error we would observe in \textit{the setting of no scenario deviation}.
\end{itemize}

Because decisions will be based on projections in 
the modeled counterfactual scenarios specified, we argue that model miscalibration is the most relevant
source of error and the best measure of whether 
underlying models are performing as intended. That is, we care most about the difference between the projected value and the value that would have been observed in the scenarios that were modeled, i.e., $e^{m}(x_i) = \mathcal{P}^{m}(y|x_i) - \mathcal{P}^{*}(y|x_i)$. To make these comparisons, though, half of the necessary information will always be missing. 

For example, information is missing even if we aim to calculate error in the scenario that was realized, $e^{m}(x^*) = \mathcal{P}^{m}(y|x^*) - \mathcal{P}^{*}(y|x^*)$, the projected value $\mathcal{P}^{m}(y|x^*)$ is likely not available in the original set of projections. It may be possible to retrospectively obtain this value by rerunning the models at the realized scenario, $x^*$, or by estimating the projected value for the scenario (e.g., through a surrogate model of the relationship between scenario projections and $x$).

The same problem arises when calculating the error that we are ultimately interested in: error at the scenarios that were modeled, $e^{m}(x^*) = \mathcal{P}^{m}(y|x_i) - \mathcal{P}^{*}(y|x_i)$. In this case, we do not know the observed value, $\mathcal{P}^{*}(y|x_i)$, and unlike models, the real world cannot be ``rerun" for different scenarios. Because error at the realized scenario can be more easily obtained, we might wonder if it could provide a proxy for error at the scenarios that were modeled. But this is not the case; different models can produce projections with the same error at the realized scenario, but with very different error at the projected scenarios (\cref{fig:base-plot}A vs. B). So, the best option is to estimate the error at the scenarios that were modeled. 

\section{Overcoming the challenge: Three approaches to estimate error for counterfactual scenario projections}
\label{S:3}

Here we outline three approaches to evaluating scenario projections. Each approach has its own requirements, assumptions, benefits, and limitations (\cref{tab:approaches}).

\begin{table}
    \centering
    \begin{tabular}{|l|p{4.5in}|}\hline 
    \multicolumn{2}{|c|}{\textbf{Approach 1:} Evaluate only plausible scenarios} \\
    \hline
    \hline
    \textbf{Requirements} & Some scenarios must be close to how the world unfolded. \\ 
    \hline
    \textbf{Assumptions} &  Must define a  ``plausible" scenario, or how close a scenario must be to reality. Assumes that any remaining scenario deviation will not affect projected outcomes.\\
    \hline
    \textbf{Benefits} & Error can be calculated for individual projections. No extrapolation is required. \\ 
    \hline
    \textbf{Limitations} & Improperly defining ``plausible” could cause misleading results. Can only evaluate a subset of projections, and the set of projections may not have any plausible scenarios.\\
    \hline
    \hline 
    %%%%%
    \multicolumn{2}{|c|}{\textbf{Approach 2:} Infer error distribution directly} \\
    \hline
    \hline
    \textbf{Requirements} & Multiple projections, with naturally occurring variation in realized scenario value. \\ 
    \hline
    \textbf{Assumptions} &  Realized scenario value is sampled independently along the scenario axis and independently from model error. Assumes models making projections capture important properties of the system. \\
    \hline
    \textbf{Benefits} & Error can be estimated for individual projections from all scenarios. \\
    \hline
    \textbf{Limitations} & Honest reprojection at the realized scenario is necessary. Statistical fitting of errors may be complicated, and could imply different error distributions across models.\\
    \hline 
    \hline
    %%%%%
    \multicolumn{2}{|c|}{\textbf{Approach 3:} Estimate observations in modeled scenarios} \\
    \hline
    \hline
    \textbf{Requirements} & Multiple projections, with naturally occurring variation in realized scenario value.  \\ 
    \hline
    \textbf{Assumptions} & Requires assumptions about relationship between scenario axis and projection axis, as well as understanding of important covariates. \\
    \hline
    \textbf{Benefits} & Error can be estimated for individual projections from all scenarios. Causal inference methods can be leveraged. \\ 
    \hline
    \textbf{Limitations} & Assumptions required may be too strong, or may miss key properties of the system. Biases in estimates of observations can bias evaluation. models.\\
    \hline
    \hline
    \end{tabular}
    \caption{Evaluation approaches. Requirements, assumptions, benefits and limitations for each evaluation approach}
    \label{tab:approaches}
\end{table}

\subsection{Approach 1: Evaluate only plausible scenarios}
\begin{figure}[!htbp]
    \centering
    \includegraphics[width=1\linewidth]{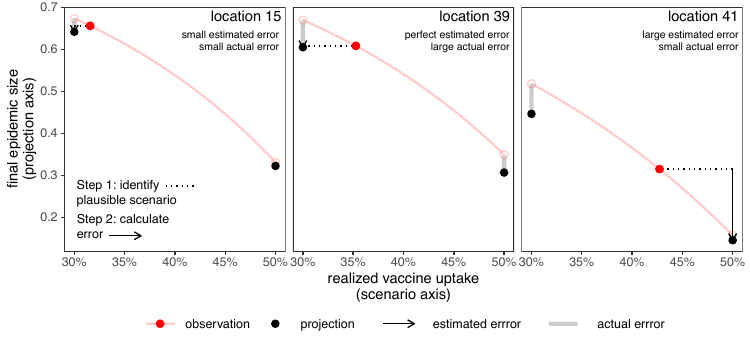}
    \caption{Overview of Approach 1. For three sample locations (panels) from a single model in the simulation experiment, projected values (black circle), observations (filled red circle), and what the observations would have been in the scenarios that were modeled (open red circle). The ``true" relationship between vaccine uptake (scenario axis) and final epidemic size (projection axis) is shown with a light red line. The true error is shown with a light gray line. Implementing Approach 1 requires two main steps. First, the plausible scenarios were identified for each location, by comparing scenario assumed vaccine uptake and realized vaccine uptake (dotted lines, step 1). In locations 15 and 29, the low vaccination scenario was retrospectively deemed plausible, and in location 41, the high vaccination scenario was plausible. Then, error is calculated by comparing the observed final epidemic size to the projected epidemic size (arrow, step 2) for each location. }
    \label{fig:approach1}
\end{figure}

The first approach treats projections as conditional forecasts by only evaluating a subset of projections for which scenarios are retrospectively deemed ``plausible" (\cref{fig:approach1}). A plausible scenario is one that can be considered sufficiently close to how the world actually unfolded, such that direct comparison between projection and observation is meaningful. For example, plausible scenarios could be defined as $x_P = \{i: |x_i-x^*|< \tau\}$ based on some threshold, although other definitions are also possible. This approach assumes that any remaining scenario deviation will not meaningfully affect projected outcomes, and that additional scenario deviation in less plausible scenarios means they cannot be directly evaluated against the observation.

For example, if we are evaluating projections of future hospitalizations conditional on different assumed vaccine uptake levels, the set of plausible scenarios would be defined retrospectively by comparing scenario-assumed vaccine uptake to realized vaccine uptake. Then projected hospitalizations in the plausible scenarios would be directly compared to observed hospitalizations \cite{howerton_evaluation_2023}. 

There are two primary challenges with this approach. First, it is not clear how close a scenario must be to reality before it is sufficiently ``plausible”. A lenient definition of plausibility risks attributing error caused by scenario deviation to error caused by model miscalibration. A strict definition of plausible may leave no scenarios left to be evaluated. Second, the results can only tell us at best about the calibration of the subset of projections that were considered to be plausible, and it may not be reasonable to assume this error is representative of error in the other scenarios that were modeled.

\subsection{Approach 2: Infer the error distribution directly}
\begin{figure}[!htbp]
    \centering
    \includegraphics[width=1\linewidth]{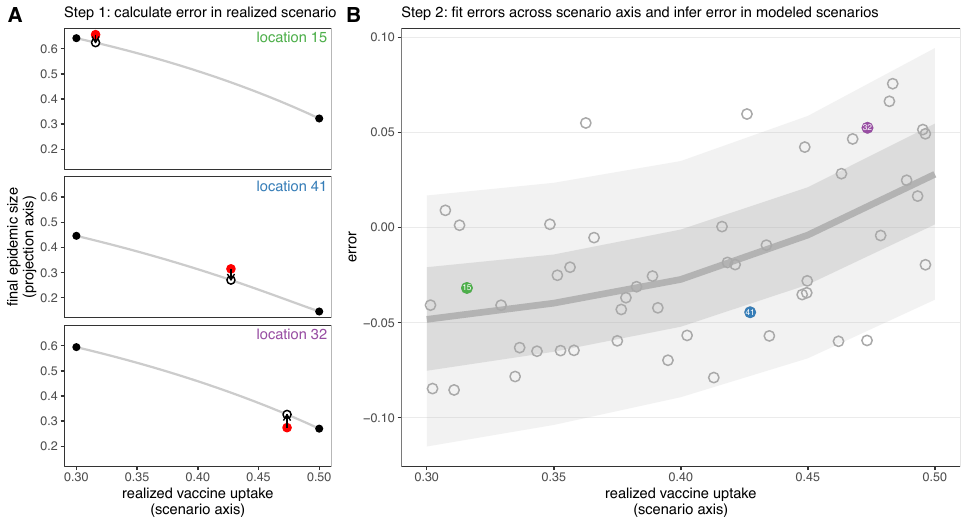}
    \caption{Overview of Approach 2. (A) For each location, error in the realized scenario is calculated by subtracting the observation (red circle) from the projection that would have been made (open black circle). Projected values for each scenario (black circles) and the projected relationship (gray line) are shown for reference. (B) Error in the realized scenario is plotted as a function of realized vaccine uptake  for each location (open gray circles), and a model is fit to estimate infer error in the modeled scenarios. The dark and light gray ribbons show the 50\% and 90\% and prediction intervals from a generalized addative model fit with a spline term for realized vaccine uptake. Three locations from the simulation experiment are highlighted as examples.}
    \label{fig:approach2}
\end{figure}

The second approach directly infers the distribution of errors across an entire set of projections (\cref{fig:approach2}). When naturally occurring variation exists in the realized scenario and projected outcome (e.g., across locations), this variation can be leveraged to model the errors in scenarios that were not realized. 

This process involves three steps. First, each model, $m$,  must be rerun based on the realized scenario conditions to obtain the value that would have been projected in the realized scenario, $\mathcal{P}^m(y|x^*)$. This projected value can then be used to calculate the error in the realized scenario, $e^m(x^*) = \mathcal{P}^m(y|x^*) - \mathcal{P}^*(y|x^*)$. Note, this error is specific to a particular model, and is calculated for every projection in the set (e.g., across locations). Then, these errors can be used to fit the relationship between error and $x$, i.e., $e^m(x) = g(x)+ \epsilon$. Finally, this fitted relationship can be used to estimate the distribution of errors across the set of projections at each assumed scenario value $x_i$.

For example, when evaluating projections of future hospitalizations in each US state across different vaccination uptake assumptions, error at the realized scenario would be calculated for each state and used to infer the distribution of errors across states for each of the scenarios that were modeled.

One main limitation of this approach is that it requires honest retrospective reprojection at each realized value $x^*$ for all models. Statistically modeling a potentially complex underlying error distribution also poses potential challenges. An error distribution for each location could be obtained if the inference is conditioned on location level covariates. Finally, although the observations are not directly estimated with this method, they are implied from the inference of error. If errors are estimated independently for each model, this could lead to the somewhat non-nonsensical situation where each models error distribution implies a different distribution of ``observations" in the counterfactual worlds. 

\subsection{Approach 3: Estimate observations in modeled scenarios}
\begin{figure}[!htbp]
    \centering
    \includegraphics[width=1\linewidth]{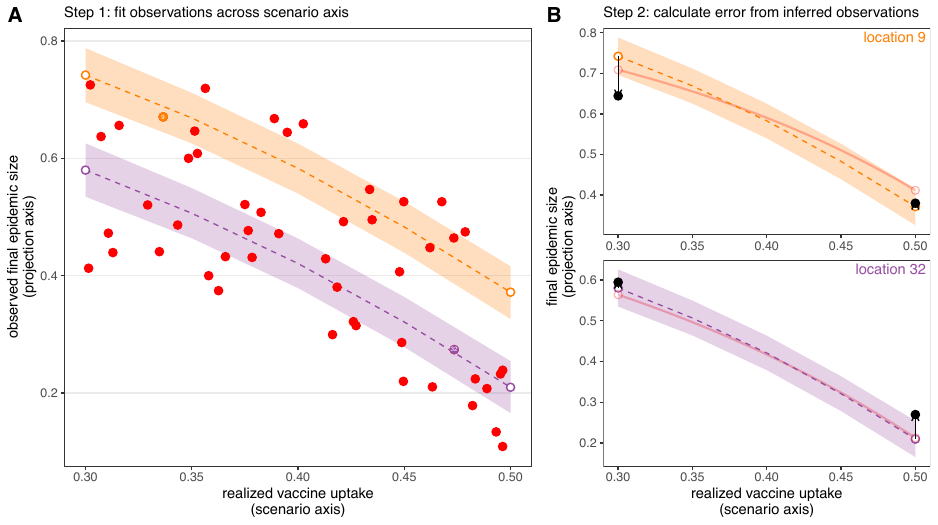}
    \caption{Overview of Approach 3. (A) Observations for each location (red circles) are fit across the scenario axis using a generalized additive model with a a spline term for realized vaccine uptake and including location-specific $R_0$ as a covariate. From this model, location-specific estimates of observations in the modeled scenarios can be made (two locations shown as examples, ribbon shows the 50\% prediction interval and line shows the median prediction). (B) Then, for each location, projected values (black circles) are compared to the inferred observations (arrows) to calculate error for each scenario that was modeled.}
    \label{fig:approach3}
\end{figure}

The third approach estimates what observations would have been in the scenarios that were modeled had their assumptions been realized (\cref{fig:approach3}). Similar to Approach 2, this approach leverages naturally occurring variation across realized worlds, but in this case, the goal is to estimate the observations and then calculate error (as opposed to inferring the errors directly). In other words, one could fit $\mathcal{P}^*(y|x) = f(x)+ \epsilon$ using observations $(x^*, \mathcal{P}^*(y|x^*))$ and the fitted model could infer $P^*(y|x_i)$.

For example, data on the relationship between vaccine coverage and hospitalizations could be fit across multiple US states using a regression model. This model could then be used to project the expected hospitalization burden in each state given the scenario specified vaccine coverage. This number can then be compared to the scenario projections to estimate calibration error (accounting for the uncertainty in both approaches). This is similar to an approach used to assess the quality of temperature projections from climate models \cite{hausfather_evaluating_2020}.

Estimating the ``observed" value for each scenario requires a range of assumptions, including assumptions about the “true" relationship between the scenario variable and the projected values as well as about the role of additional covariates in changing projected outcomes. These assumptions will be based on knowledge of the system, and there are well-established methods for making such inferences from the field of causal inference \cite{hernan_estimating_2006}. Nevertheless, the necessary assumptions may not always be possible, and biases in estimates of the observations could bias estimates of calibration. In particular, this approach limits you to evaluation based on observations for which such surrogate models could be easily fit (e.g., total hospitalizations across a season), though other outcomes may be of primary interest (e.g., hospitalizations for each week of the season). 

\subsection{Strengths and weakness of each approach}
Based on the theoretical framework that we have presented, we can see that obtaining error at the scenarios that were modeled will always require extrapolation to unseen conditions. Choosing the right approach will help minimize biases in that extrapolation. Here we compare the approaches and discuss the situations in which each is most suitable (\cref{tab:approaches}). 

While the first approach does not require extrapolation of observations or error, it does require strong assumptions for the results to be interpretable. Differences between scenario and reality are assumed not to be important, which is unlikely in scenario modeling contexts. Moreover, the approach can only provide information about the calibration of projections for scenarios that were close to what materialized; hence, we will not have meaningful error estimates for the other modeled scenarios (as discussed in \cref{S:2} and shown in \cref{fig:base-plot}). Given these limitations, we recommend Approach 1 only when the other approaches are not possible, and even then, it should be interpreted with caution.

The second and third approaches offer meaningful results for all scenarios, but they require stricter assumptions. Strong statistical techniques and validation are essential. If a good model of the observations can be created, Approach 3 should be favored. Approach 3 can leverage established methods from causal inference, which use available covariates and reasonable, verifiable assumptions to make inferences about observations in scenarios that were not realized. 

Compared to the third approach, estimating a distribution of errors across projections for each model (as in Approach 2) has three main drawbacks. First, the implied distribution of observations (which can be obtained from the projected values and the distribution of errors) may be different across models if error is estimated independently for each model. Second, to calculate error for every projection that was made (e.g., for every location), location-specific covariates are required, but identifying predictors of error is less straightforward. Third, and most importantly, this approach requires retrospectively generating projections at the realized scenario value. This may be difficult in some modeling frameworks, and requires honest re-projectors that do not use knowledge gained since projections were made (when there may be incentives to do so). The last of these could be mitigated by the \textit{a priori} development of proxy models that can provide expected model projections at any scenario value. 

Yet, in situations where it is not possible to create a good statistical model of observations, the second approach (inferring errors directly) may be favorable. In particular, the models generating projections may capture important non-linear relationships in the system that a statistical model cannot. For example, herd immunity is known to be an important feature of infectious disease outcomes as vaccination coverage increases. Mechanistic models that generate projections may capture these non-linearities better than a statistical model. Ultimately, comparing the results of both approaches is likely the most robust strategy.

\section{Separating error into components driven by scenario deviation and model miscalibration
}
Each of these approaches are designed to isolate error that arises from model miscalibration without any influence from scenario deviation. A good estimate of this error also allows us to show how much of the observed deviation that we see when comparing a projection to an observation directly is driven by the scenario not matching reality. This possibility comes from decomposing observed deviation into two linear components, representing model calibration error and scenario specification error: 

\begin{align*}
P^m(y|x_i) - P^*(y|x^*) & =P^m(y|x_i) - P^*(y|x_i)+P^*(y|x_i)-P^*(y|x^*) \\
\text{observed deviation} &  = \text{model calibration error}+\text{scenario specification error}
\end{align*}

\begin{figure}[!htbp]
    \centering
    \includegraphics[width=0.75\linewidth]{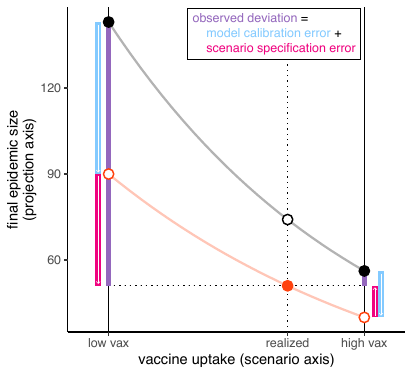}
    \caption{Example of error decomposition. Total error is defined as the difference between projection and observation (purple, $P^m(y|x_i) - P^*(y|x^*)$), and can be separated into two components: model calibration error (blue, $P^m(y|x_i) - P^*(y|x_i)$ and scenario specification error (pink, $P^*(y|x_i)-P^*(y|x^*)$).}
    \label{fig:decompose-error}
\end{figure}

The most important observation that follows from this simple formulation is that the two sources of error can act in different ways (\cref{fig:decompose-error}). In the intuitive case, model calibration error and scenario specification error are additive, leading to observed deviation that is larger than either component (low vaccination scenario, \cref{fig:decompose-error}). However, there is also the possibility for the less intuitive case where the errors are in opposing directions, leading to a observed deviation that is smaller than either component (high vaccination scenario, \cref{fig:decompose-error}). In this case the model ``gets the right answer, but for the wrong reason”. 
Hence, we argue that there is a theoretical ``total error" that is different from the observed deviation and is sum of the two absolute errors:

\begin{align*}
\text{total error} & =|P^m(y|x_i) - P^*(y|x_i)|+|P^*(y|x_i)-P^*(y|x^*)| \\
\end{align*}

\noindent
This error provides a more complete summary of the two components. 

\section{Demonstrating each approach via simulation experiment}
\label{S:4}

We use a simulation experiment to demonstrate and test each of the three approaches. Unlike the real world, the simulation experiment allows us to know what would have been observed in the scenarios that were modeled but not realized, as well as in a ``realized" scenario. Thus, we can assess how well each approach estimates the errors of interest.

\subsection{Simulation setup}

We imagined a case where scenario projections are requested for the number of infections expected at the end of an outbreak (epidemic size) given low and high vaccination coverage scenarios. Projections were needed for 50 locations. A summary of the simulation setup is provided in \cref{tab:simsetup}, and each step is described below. Additional details of the simulation experiment, including model equations, are provided in the Supplementary Material. 

\subsubsection{The ``True” Model}
We assumed that the true generating process was an Susceptible-Infected-Recovered (SIR) model from epidemiology. The final size of an epidemic depends on (1) how quickly the pathogen spreads, which is characterized by the basic reproduction number, $R_0$, (2) how many individuals have been vaccinated and are protected from infection, $v$, and (3) assumptions about how heterogeneities in population structure affect transmission, which we accounted for with an exponent on the infection term, $\alpha$. Larger $R_0$, lower $v$, and higher $\alpha$ all lead to larger outbreaks. We randomly drew a ``true” $R_0$ value, $R_{0,l} ^*  $ and $\alpha$, $\alpha^*_l$ for each location. 

\subsubsection{Projection Models}
Each projection model was also an SIR model that deviated from the true model in the following ways. First, we assumed that each model imperfectly estimated the ``true" location-specific basic reproduction number, $R_{0,l}^*$. In particular, we assumed that each model had an overall bias that holds across locations, $b_m$ (systematically over- or underestimating), and has an additional location-specific bias, $b_l^m$. Together, these biases amount to the $R_0$ ``estimated” by each model for each location, $R_{0,l}^m =R_{0,l}^* + b^m+b_l^m$. We randomly drew each of these values. Similarly, we randomly drew model-specific $\alpha^m$ and introduced additional variation across locations, $\alpha_l^m$.

\subsubsection{Scenario Projections of Epidemic Size}
Scenario projections were made from each of the projection models at values of 30\% and 50\% vaccination coverage. This yielded a projection from each model for both scenarios in each of the 50 locations, $\mathcal{P}_l^m(y|x_i)=SIR(R_{0,l}^m, \alpha_l^m,x_i)$. 

\subsubsection{``Observed” Vaccination Coverage and Epidemic Sizes}
So that the realized scenario value did not match the modeled scenario values, we also generated a realized vaccination coverage value for each location $x_l^* $. Then, we calculated the observed epidemic size for each location from an SIR model with the realized vaccination coverage and the ``true” location-specific parameters,  $\mathcal{P}_l^*(y|x_l^*)=SIR(R_{0,l}^*, \alpha_l^*,x_l^*)$.

\subsubsection{Epidemic Sizes in Counterfactual Worlds}
\label{sect:espi_sizes_counterfact}
In addition to observed epidemic sizes, we also calculated what the observation would have been in the 30\% and 50\% vaccination coverage scenarios using the ``true” model. This yielded $\mathcal{P}_l^*(y|x_i)=SIR(R_{0,l}^*, \alpha_l^*,x_i)$ for $x_i \in \{30\%, 50\%\}$. A perfect model would make the projection $P_l^m(y|x_i) = P_l^*(y|x_i)$. 

\subsubsection{Projected Epidemic Size in the Realized Scenario}
Finally, we reran the projection models at the observed vaccination coverage value to generate what the models would have projected in the realized scenario, $P_l^m(y|x_l^*)=SIR(R_{0,l}^m, \alpha_l^m,x_l^*)$. In this case, we assumed that the value at the observed scenario value was reprojected honestly and exactly.

\begin{table}
    \centering
    \begin{tabular}{|p{1.9in}|p{2.1in}|p{2.1in}|}
    \hline
         & \textbf{``True" model} & \textbf{``Projection” models}\\
         \hline
         \textbf{Vaccination coverage
} & $x_l^* \sim U(0.3, 0.5)$ & $x_1 = 30\%$, $x_2 = 50\%$\\
(scenario values) && \\
\hline
         \textbf{Basic reproduction number} & $R_{0,l}^* \sim U(2,3)$ & $R_{0,l}^m =R_{0,l}^* + b^m+b_l^m$
where $b^m \sim N(0, 0.05)$ and \\
(intermediate value) & & $b_l^m \sim N(0, 0.05)$ \\
\hline
         \textbf{Heterogeneity constant} & $\alpha_l^* \sim N(0.975, 0.01)$ & $\alpha_l^m \sim U(m, 0.01)$ with \\
         (intermediate value) & & $m \sim U(0.95, 1)$\\
\hline
         \textbf{Epidemic size} & $\mathcal{P}_l^*(y|x_i)= SIR(R_{0,l}^*,\alpha_l^* ,x_i)$
 & $\mathcal{P}_l^m(y|x_i) = SIR(R_{0,l}^m,\alpha_l^m, x_i)$\\ 
(projected outcome) &  $\mathcal{P}_l^*(y|x_l^*)=SIR(R_{0,l}^*,\alpha_l^* , x_l^*)$& $\mathcal{P}_l^m(y| x_l^*)=SIR(R_{0,l}^m ,\alpha_l^m, x_l^*)$\\
\hline
    \end{tabular}
    \caption{Overview of simulation setup. The simulation experiment generated both scenario values ($x$ values) and projected outcomes ($y$ values) from a ``true” model and a range of imperfect ``projection” models. The ``truth” is denoted with a star (e.g., $x^*$ and $\mathcal{P}^*(y|x))$ and individual models are denoted with an $m$ (e.g., $P^m(y|x)$); $l$ denotes individual locations and $SIR(R_0, \alpha, v)$ represents the total number of infections resulting from simulating an SIR model with basic reproduction number $R_0$, vaccination level $v$, and heterogeneity constant $\alpha$. Details on the SIR model and implementation are provided in the Supplementary Material.
}
    \label{tab:simsetup}
\end{table}

\subsection{Implementing each approach to estimate error}
Using the simulated values, we implemented each of the three approaches to error estimation defined above (\cref{S:3}). Ultimately, this yielded an estimate of the distribution of errors across locations for a single model (and, in some cases, an estimate of error for each location and model).

\subsubsection{Implementing Approach 1: Finding the most plausible scenario}
To define the most plausible scenario, we calculated the absolute difference between the realized vaccination coverage and the scenarios that were modeled, i.e., $|x_l^*-0.3|$ or $|x_l^*-0.5|$. We defined the most plausible scenario, $x_l^P$  as the scenario for which this value was minimized (\cref{fig:approach1}). Then, we calculated the error between the projected value in the most plausible scenario and the observation, $\mathcal{P}_l^m(y|x_l^P) - \mathcal{P}_l^*(y|x_l^*)$. The distribution of errors for a given scenario was calculated across only those locations for which that scenario was most plausible.

\subsubsection{Implementing Approach 2: Inferring error distributions directly}
To infer the error distributions directly, we first calculated the error in the realized scenario for each location, $e_l^m(x^*)= P_l^m(y|x^*) - P^*(y|x^*)$. Then, we fit these errors with a statistical generalized additive model (GAM), including a cubic spline term for observed vaccination coverage (\cref{fig:approach2}). To test the importance of location-specific covariates, we fit this model with and without a linear term for location-specific $R_0$. Although in practice the true $R_0$ for each location would not be known when performing this estimation, this provides a ``best case scenario” in terms of the information available for fitting the model. We then inferred the distribution of errors at each of the scenarios that were modeled (across all locations, or for each location individually, depending on whether covariates were included). 

\subsubsection{Implementing Approach 3: Estimating observations in vaccination scenarios}
To infer observations in the vaccination scenarios, we fit a GAM to the observed final size values across locations with a cubic spline term for observed vaccination coverage (\cref{fig:approach3}). To match the methods for Approach 2, we also fit this model with and without a linear term for location-specific $R_0$. 

We then extrapolated the fitted model to each of the scenarios that were modeled (i.e., $x_i \in \{0.3,0.5\}$). When location-specific covariates were included, we obtained an estimate of the observation in each location and when there were no location-specific covariates, we obtained a distribution of observations across locations. 

Finally, we calculated error using the estimates of what the observations would have been. We randomly drew 10,000 samples from the distribution of possible observations and calculated error for each of these samples, yielding a distribution of possible errors. When we had one distribution for each location, we pooled all samples and summarized across locations for the final error distribution.

\subsubsection{Comparing estimated errors to actual errors}
We assessed the accuracy of each approach by its ability to recreate the true distribution of errors, in terms of the mean and the full distribution. Because we had specified a “true” set of parameters, we were able to generate what the observations would have been in the scenarios that were modeled (see \cref{sect:espi_sizes_counterfact}) and we used this to calculate the ``true” model error. 

We used two metrics to assess how well each approach captured the true errors. First, mean absolute error compared the absolute value of the difference between the mean of the estimated error distribution and the mean of the true error distribution (or in the case of the location-specific estimates, just the true error). Second, we used the Kolmogorov-Smirnov (KS) test with a p-value of 0.05 to determine whether the distributions were significantly different (or not).

\subsection{Simulation results}
\begin{figure}[!htbp]
    \centering
    \includegraphics[width=1\linewidth]{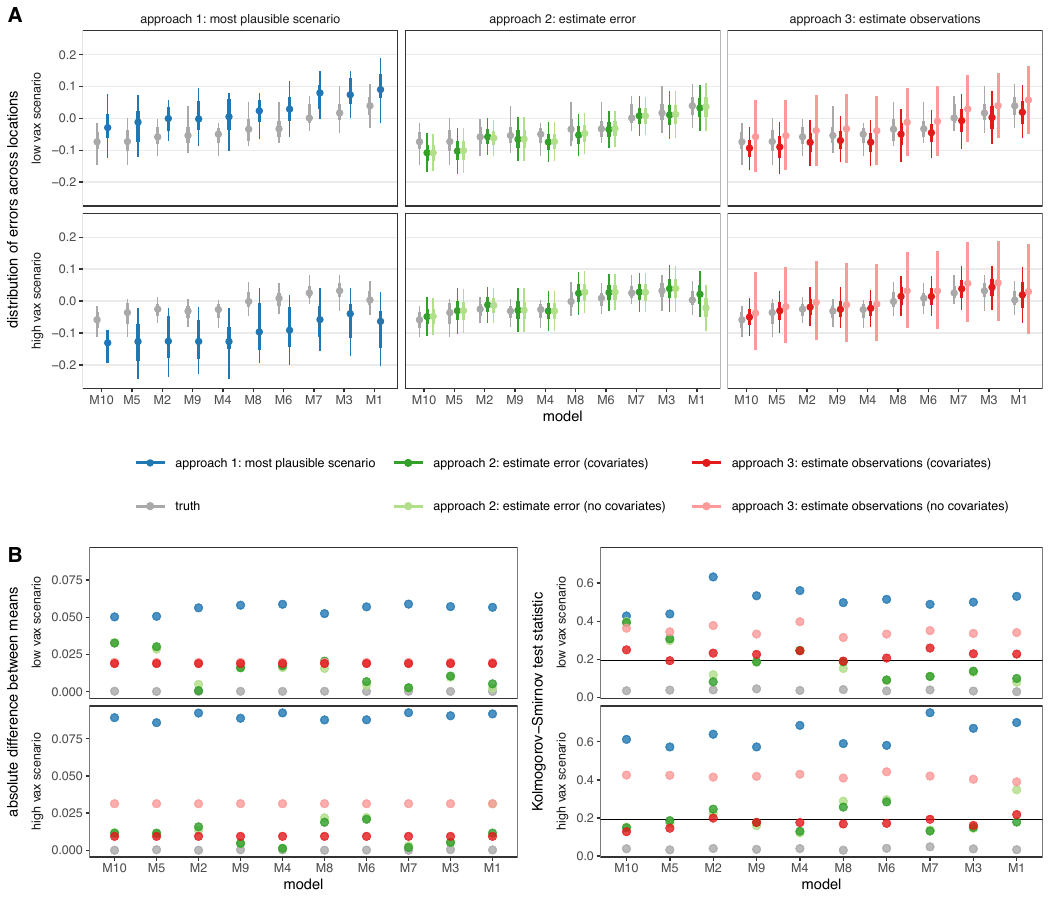}
    \caption{Summary of simulation experiment results. (A) Predicted distribution of model miscalibration errors across locations for each approach: Approach 1 (dark blue), Approach 2 with and without covariates (dark and light green, respectively), and Approach 3 with and without covariates (dark and light red, respectively) are compared to the true distribution of errors (gray). (B) Evaluation of performance for each approach, as measured by (left) the absolute difference between mean of the true distribution and mean of the estimated distribution, and (right) Komogorov-Smirnov (KS) statistic. For the KS statistic, the horizontal black line shows the value corresponding to a p-value of 5\%. Points below the black line suggest the distributions are not significantly different.
}
    \label{fig:sim-results}
\end{figure}

Our simulation experiment demonstrates the strengths and weaknesses of each approach (\cref{fig:sim-results}). Approach 1 is systematically biased in the estimation of miscalibration error for both scenarios. This arises because scenario deviation has not been completely accounted for. There are cases where scenario specification error offsets the miscalibration error and estimated error is near zero while actual error is not. There are other cases where projections were almost perfect but scenario deviation makes estimated error artificially large (\cref{fig:approach1}, left/middle panel vs. right panel). Locations with a larger difference between scenario-assumed and realized vaccine uptake have more biased estimates of error under Approach 1 (\cref{fig:mae-byscenario}). 

Approaches 2 and 3 produce estimates of the error distribution that are more accurate than Approach 1, especially when covariates are included in the estimation. These two approaches are comparably accurate in their mean estimates and capture the full distribution in 14/20 model-scenario comparisons for Approach 2 and 10/20 for Approach 3 (according to a KS test with a p-value of 5\%). Approach 3 has a slight bias in estimates of error in the low vaccination scenario due to overestimation of observations in this scenario for high $R_0$ locations. Approach 2, where models capture the non-linearities of the SIR model, does not have this problem as consistently. 

It is worth noting, however, that bias in estimates of observations affects estimates of error for all models equally for Approach 3. This is illustrated by the equal difference between true and estimated means for all models (red points, \cref{fig:sim-results} panel A). This is not the case in Approach 2, and thus results are less consistent across models. Because we fit each model independently in this approach, the inferred observations can vary across models, but in this case, the inferred distribution of observations is not significantly different from the true distribution (\cref{fig:infered_error_across_locs}).

Finally, the inclusion of covariates enables the generation of location-specific estimates of error (\cref{fig:eg_error_ests}). Including covariates does not have a significant effect on the performance of Approach 2, but for Approach 3, covariates are essential (\cref{fig:sim-results}). Without covariates, the variance of the inferred error distribution is too large as only a distribution of observations across locations can be inferred and used to calculate error. When assessing the performance of each approach in capturing location-specific errors (rather than the distribution of errors across locations), Approaches 2 and 3 were comparable, with Approach 2 better capturing locations with a higher $R_0$ which have the strongest non-linear effects (\cref{fig:mae_loc_specific,fig:mae_a2_a3}). 

\section{Discussion}
Here we present a framework and specific methods for retrospective evaluation of projections from scenario modeling exercises. We illustrate how observed error arises from a combination of model miscalibration and differences between scenario assumptions and reality (scenario deviation). We argue that isolating the error due to model miscalibration in the counterfactual scenarios is a critical component of appropriate evaluation of scenario projections, and describe and contrast three approaches to doing so. The ability to use these approaches is not independent of scenario design. 

Each method has strengths and weaknesses, but in most cases we believe Approach 3, based on estimating observations in counterfactual scenarios, will be most appropriate. 
%The advantages of this approach are that (1) it leverages well established causal inference techniques, (2) it does not require rerunning models (honestly) under new conditions, and (3) it provides a clear path to estimating location specific errors. 
However, in cases where models capture drivers of outcomes that are not easily captured in a statistical model (e.g., due to non-linear effects related to a known mechanism), directly estimating the error distribution as in Approach 2 may work better. When possible, doing both may provide the most robust characterization of error. 

We believe there are few cases where Approach 1 should be used. The limited exceptions that do exist (e.g., evaluation of an axis with no clear quantitative ordering) may be avoidable with careful scenario design. Regardless, all error estimates from Approach 1 carry the caveat that they include error from both scenario deviation and model miscalibration. 

An important result of the ability to evaluate scenario models in counterfactual worlds is that it allows us to partition observed deviation into error stemming from scenario design and from model miscalibration (\cref{fig:decompose-error}). The communication of this, and its importance, is easy when the two errors are in the same direction, and hence the total error is their sum. In these cases, it is straightforward to explain, for instance, that the models worked well but scenario mis-specification led to a large observed difference between projections and reality. More difficult is explaining those cases where errors due to model miscalibration and scenario deviation cancel, leading to differences between projections and reality that are smaller than either source of error. In these cases we might be presented with the difficult task of explaining that a particular model was wildly miscalibrated despite projections that seem to almost perfectly match observations. This latter example highlights both the essential nature of rigorous evaluation methods and the need for research into how we effectively communicate the performance of scenario projection models and build trust in their results. Further, this phenomena reinforces the shortcomings of Approach 1 as a primary method of evaluation. 

The ease of evaluating scenarios is not independent of scenario design, and there are several properties that a set of scenarios must have to most effectively use the techniques presented here. First, the realized values of scenario axes must be measurable and, ideally, will be defined based on some continuous scale. Second, there must be enough diversity between groups (e.g., locations)  in realized values across a scenario axis to use statistical methods to predict the outcome (or error) at the specified counterfactual scenario values. Third, careful thought is needed in the definition of multi-axis scenarios to ensure that it is possible to model the simultaneous effect of both scenario axes on outcomes (or error). Finally, there are many other assumptions required when building scenario models (e.g., a common value of vaccine effectiveness across scenarios), and careful consideration of how these will be treated during evaluation is important. In one sense, such assumptions represent a scenario axis on their own, albeit one with a single value. In this case, the previous considerations remain important for these assumptions. Alternatively, the assumptions could be considered as part of the modeling process, and any resulting error would be considered a component of model miscalibration. We recommend specifying an evaluation plan during the scenario design stage to increase the transparency of these decisions and guard against the possibility of using them to artificially improve performance assessments. 

The analysis presented here has several limitations. First, we used a relatively simple simulation framework for testing the validity of our methods. Real systems are far more complex, and may therefore have complex observation and error distributions that present challenges for our methods in practice. However, any valid approach to counterfactual evaluation should work in our toy worlds. Second, we assume that the underlying structure of the models matches the true data generating process relatively well. While we do have some deviation in structure (i.e., the $\alpha$ term in the true model), no fundamental component of the generating process is missed. Third, we only assess the error distribution of point predictions, whereas most scenario projections are probabilistic in nature. This is a key area for future work, but theoretical grounding of the approaches should remain the same. Finally, we assume a very generous data environment for assessing models, including perfect observation of covariates, measurements of outcomes, etc. Observation error will impact real world performance.

Ultimately scenarios projections are made for some scientific or practical purpose \cite{runge_scenario_2024}, and fulfilling that purpose must take priority. In some cases this means that we will have to design scenario axes we know are not evaluable. Still, evaluation is important for building trust in scenario models and pushing forward the science that underlies them. Hence, scenario evaluation should be both anticipated and performed whenever possible. Our framework highlights the essential component of interpretable scenario evaluation: assessment of calibration should be done for the counterfactual worlds in which scenarios were run. There are many other aspects of the scenario modeling process we want to evaluate, such as the value of a set of scenarios for decision making, or the performance of the scenario specifications themselves. However, an assessment of projections in counterfactual worlds is a necessary step for all these assessments. This is a difficult task, but a necessary one.

\section{Acknowledgments}
There are many people we would like to thank for thoughtful discussion and feedback on this project, including members of the US Scenario Modeling Hub, Stephen Cole, Jess Edwards, Michael Hudgens, and Paul Zivich, and finally, the Isaac Newton Institute for Mathematical Sciences, Cambridge, for support and hospitality during the programme Modelling and Inference for Pandemic Preparedness where work on this paper was undertaken. We espeically thank workshop participants John Edmonds, Carl Pearson, Lilith Whittles, and Cecile Viboud. This work was supported by EPSRC grant no EP/R014604/1. In addition, EH has been funded in whole or in part with Federal funds from the National Cancer Institute, National Institutes of Health, under Prime Contract No. 75N91019D00024, Task Order No. 75N91023F00016. The content of this publication does not necessarily reflect the views or policies of the Department of Health and Human Services, nor does mention of trade names, commercial products or organizations imply endorsement by the U.S. Government. 

\section{Code Availability}
All code necessary to reproduce the figures and simulation experiment can be found at \url{https://github.com/UNCIDD/scenario-eval-theory}.



\bibliography{refs.bib}

%%%% figures

\appendix
\section{Supplementary Material}
\label{S.6}

\renewcommand\thefigure{S\arabic{figure}}    
\setcounter{figure}{0}    

\subsection{SIR setup for simulation experiment}
In the simulation experiment, we use an Susceptible-Infected-Recovered model from epidemiology. The SIR model is defined by a system of ordinary differential equations: 

\begin{align*}
    \frac{dS}{dt} &= -\beta S I^\alpha/N \\
    \frac{dI}{dt} &= \beta S I^\alpha/N - \gamma I \\
    \frac{dR}{dt} &= \gamma I \\
\end{align*}

\noindent
where $\beta$ is the transmission rate, $\alpha$ is the heterogeneity constant, and $\gamma$ is the recovery rate. The basic reproduction number $R_0 = \frac{\beta}{\gamma}$. For all simulations we set $\gamma = 10 \text{ days}$ and we calculated $\beta$ to yield the sampled $R_0$ value. We assumed that all vaccination, $v$, occurred before the start of the simulation, such that the initial conditions for each simulation was $S(0) = 1-v, I(0) = 0.001, R(0) = v - 0.001$. We numerically solved the system for one and a half years, using the R package \texttt{deSolve}. Finally, we reported the final epidemic size relative to the proportion of originally susceptible individuals after vaccination, $\frac{S(0)-S(\infty)}{S(0)}$ where $S(\infty)$ is the number of individuals in the susceptible class at the final time step. 

\subsection{Supplementary Figures}

\begin{figure}[hbt!]
    \centering
    \includegraphics[width=\linewidth]{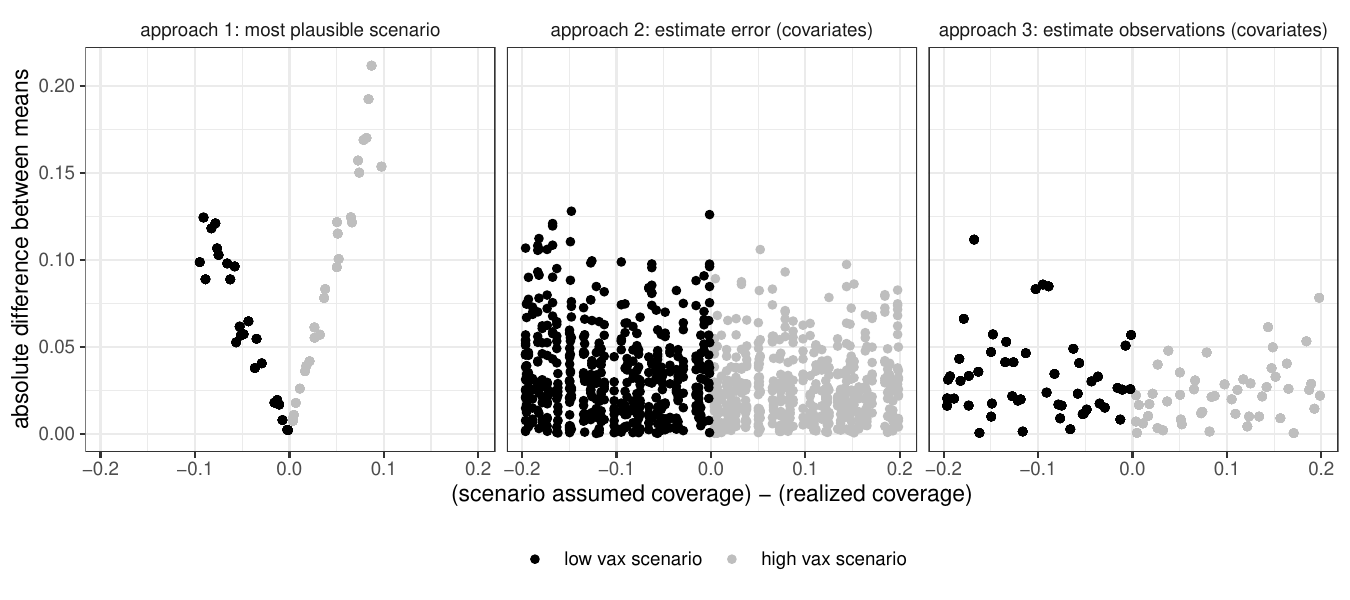}
    \caption{Performance of each approach by scenario deviation. Scenario deviation is calculated as the difference between scenario assumed vaccination coverage and and realized coverage. Results are shown for the low vaccination scenario (black) and the high vaccination scenario (gray). Each point is a single location-model pair. For Approaches 1 and 3, absolute difference between the projected mean error and the true error are equivalent across scenarios for all models (and thus there are fewer points).}
    \label{fig:mae-byscenario}
\end{figure}

\begin{figure}[hbt!]
    \centering
    \includegraphics[width=\linewidth]{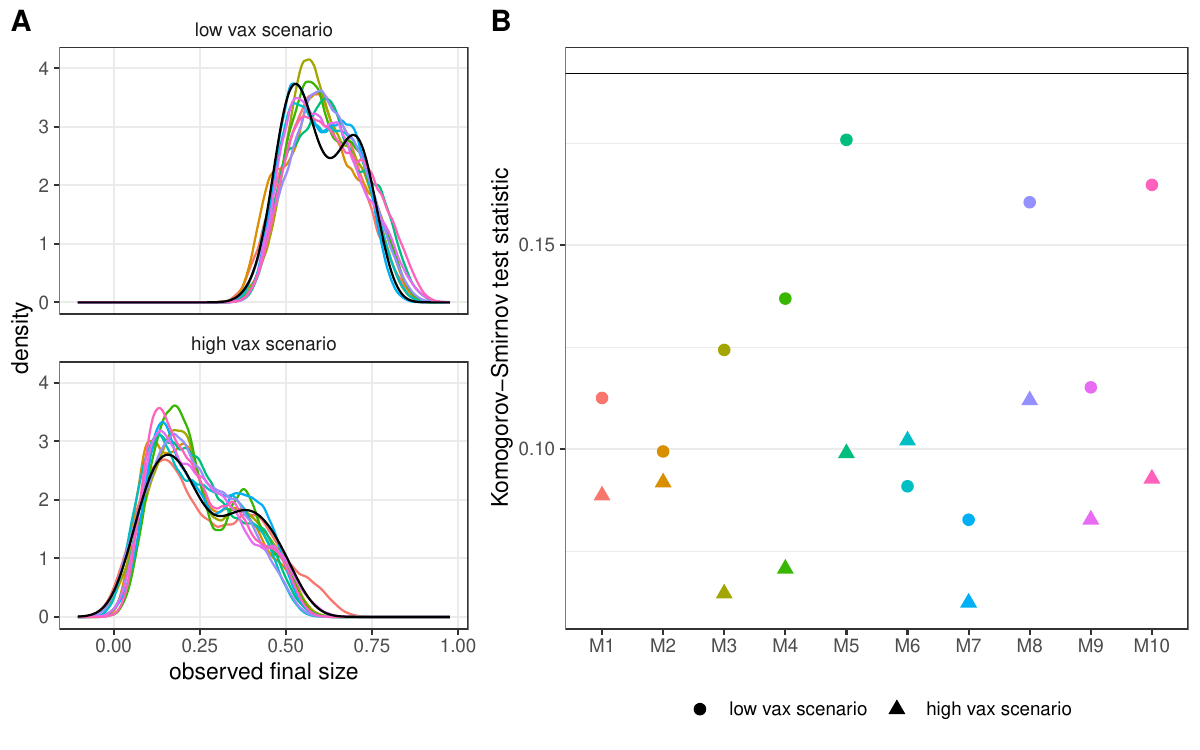}
    \caption{(A) Inferred distribution of observed final size when using Approach 2 for each model (colors) and the true distribution (black). The inferred observation distribution is calculated by taking samples from the estimated error distribution for each location and subtracting the projected value. Because model errors were estimated independently, this distribution is slightly different across models. (B) Kolmogorov-Smirnov test of difference between the inferred and true distributions in Panel A for each model and both scenarios. The horizontal black line shows the value corresponding to a p-value of 5\%. Points below the black line suggest the distributions are not significantly different.}
    \label{fig:infered_error_across_locs}
\end{figure}

\begin{figure}[hbt!]
    \centering
    \includegraphics[width=\linewidth]{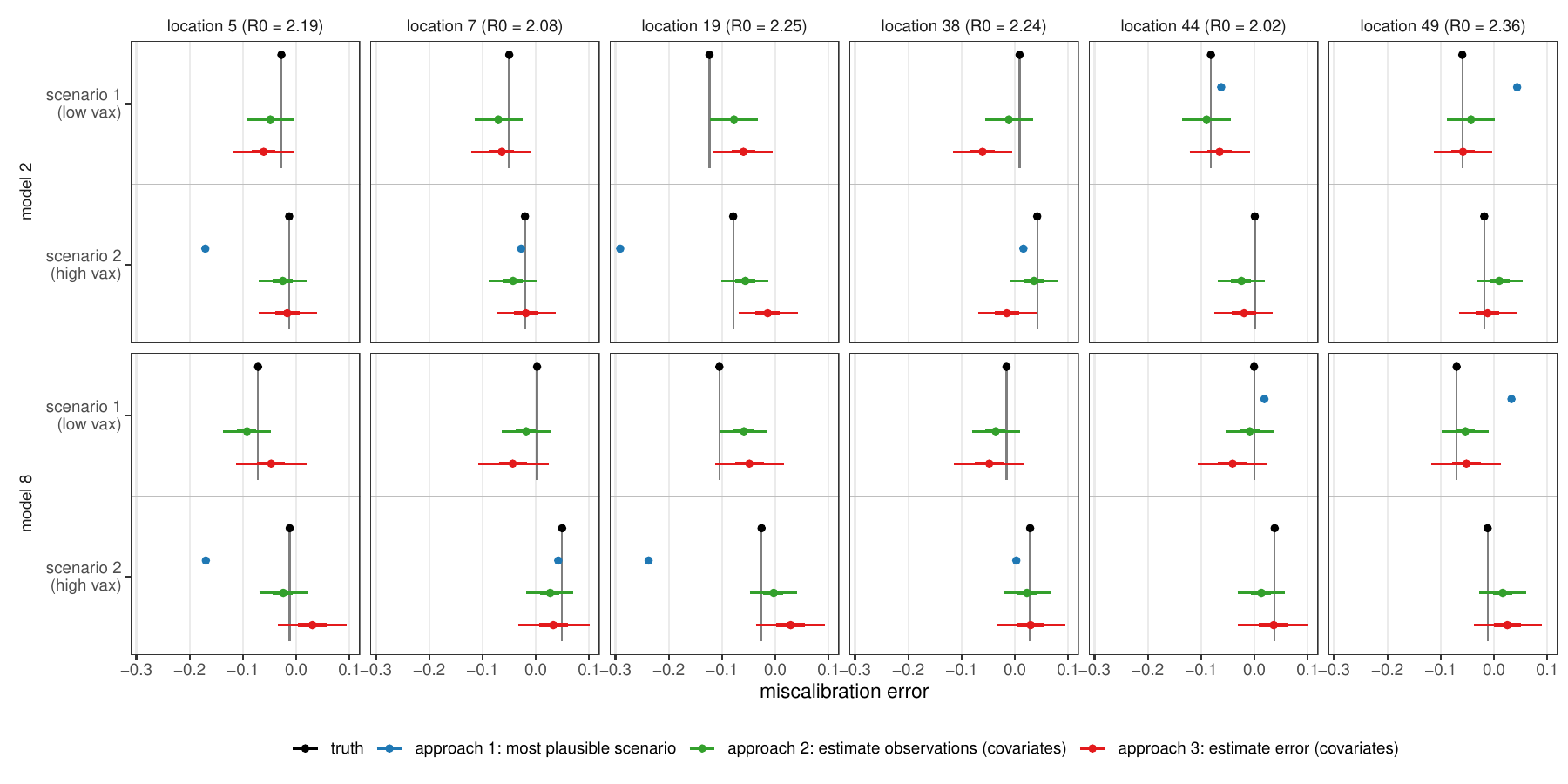}
    \caption{Examples of location-specific error distributions inferred from each approach from six randomly chosen locations and two randomly chosen models. The true error is shown as a black circle and vertical gray line for reference. Approach 1 (blue) does not generate a distribution for each location, and can only provide an estimate if the scenario is plausible. Approaches 2 (red) and 3 (green) generate a distribution of error, where the circle represents the median, the thick segment shows 50\% prediction interval, and the thin segment shows 90\% prediction interval.}
    \label{fig:eg_error_ests}
\end{figure}

\begin{figure}[hbt!]
    \centering
    \includegraphics[width=\linewidth]{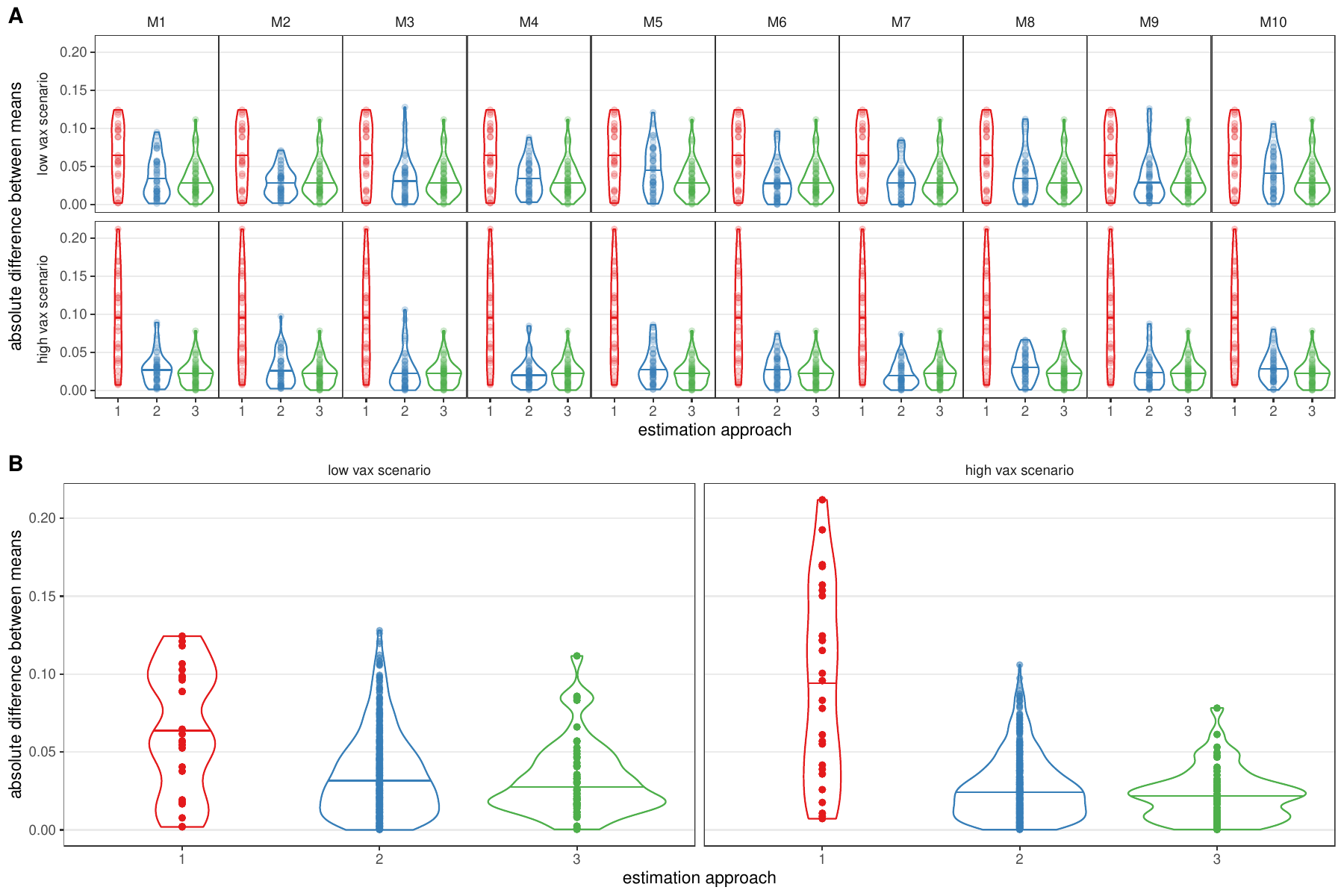}
    \caption{(A) Absolute difference between estimated error and true error across all models and scenarios. Each point represents an individual location, and the distribution across locations is summarized (where the median is represented by a horizontal line). (B) Location-specific performance summarized across all models. For Approaches 2 and 3, estimation included $R_0$ covariates. }
    \label{fig:mae_loc_specific}
\end{figure}

\begin{figure}[hbt!]
    \centering
    \includegraphics[width=\linewidth]{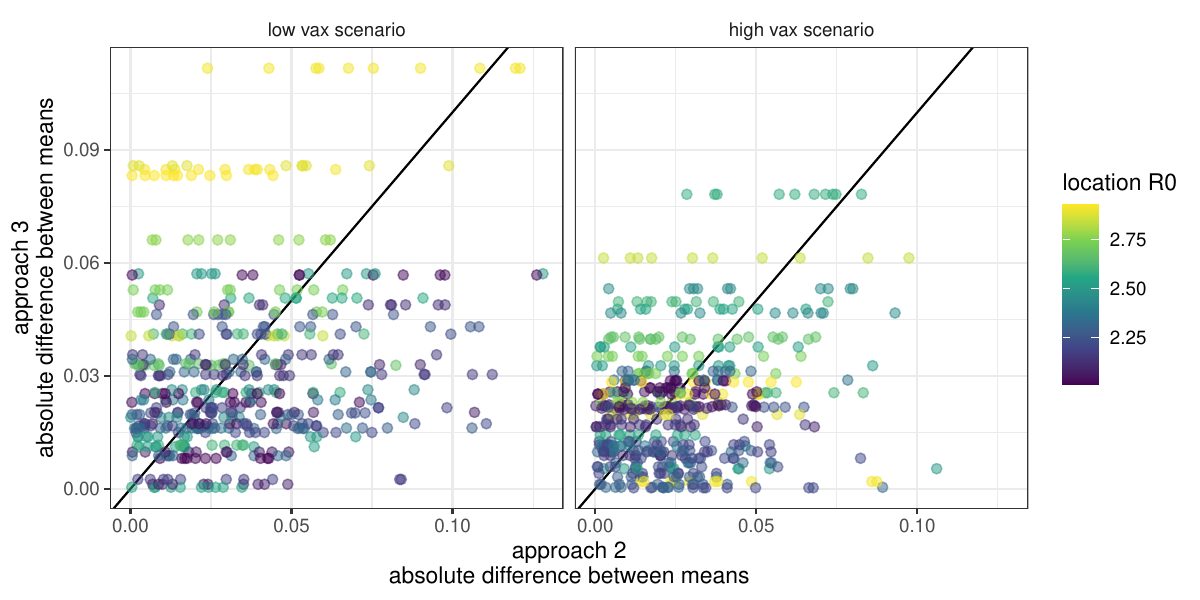}
    \caption{Absolute difference between mean error and true error in each location when estimated using Approach 2 and Approach 3. Each point represents a single location and model. The color of the point is the true location-specific $R_0$.}
    \label{fig:mae_a2_a3}
\end{figure}

\end{document}